\documentclass[a4paper,11pt]{article}
\usepackage[left=2.5cm,right=2.5cm,top=3cm,bottom=3cm]{geometry}
\usepackage[utf8]{inputenc}
\usepackage[american]{babel}
\usepackage{lineno}
\usepackage{csquotes}
\usepackage{sansmath}
\usepackage{amsmath, amssymb, amsthm, mathtools}
\usepackage{nicefrac}
\usepackage[ttscale=.875]{libertine}
\usepackage[scaled=0.96]{zi4}
\usepackage{adjustbox}
\usepackage{parskip}
\usepackage{graphicx}
\usepackage{graphbox}

\usepackage[numbers]{natbib}

\usepackage[labelsep=period]{caption}

\usepackage{xcolor}
\definecolor{blendedblue}{rgb}{0.2, 0.2, 0.6}
\definecolor{blendedred}{rgb}{0.8, 0.2, 0.2}
\usepackage[bookmarks=true,
            breaklinks=true,
            pdfborder={0 0 0},
            citecolor=blendedblue,
            colorlinks=true,
            linkcolor=blendedblue,
            urlcolor=blendedblue,
            citecolor=blendedblue,
            linktocpage=false,
            hyperindex=true,
            linkbordercolor=white]{hyperref}
\hypersetup{colorlinks=true}

\usepackage{enumitem}
\usepackage{caption}
\usepackage{subcaption}
\usepackage{algorithm}
\usepackage{algorithmic}
\usepackage{amsmath}
\usepackage{amsfonts}
\usepackage{amssymb}

\makeatletter
\newcommand{\crossout}[1]{%
  \begingroup
  \sbox\z@{#1}%
  \dimen\z@=\wd\z@
  \dimen\tw@=\ht\z@
  \dimen\z@=.99626\dimen\z@   
  \dimen\tw@=.99626\dimen\tw@ 
  \edef\co@wd{\strip@pt\dimen\z@}
  \edef\co@ht{\strip@pt\dimen\tw@}
  \leavevmode
  \rlap{\pdfliteral{q 1 J 0.4 w 0 0 m \co@wd\space \co@ht\space l S Q}}%
  \rlap{\pdfliteral{q 1 J 0.4 w 0 \co@ht\space m \co@wd\space 0 l S Q}}%
  #1%
  \endgroup
}
\makeatother

\author{Anthony Strock$^{2,1,3}$, Nicolas P. Rougier$^{1,2,3,\dagger}$ and Xavier Hinaut$^{1,2,3,\dagger,*}$}

\date{\small
$^1$Inria Bordeaux Sud-Ouest, Talence, France\\
$^2$LaBRI, Université de Bordeaux, CNRS UMR 5800, Talence, France\\
$^3$IMN, Université de Bordeaux, CNRS UMR 5293, Bordeaux, France\\
$\dagger$ Equal contribution,
$^*$Corresponding author: xavier.hinaut()inria.fr}

\title{Transfer between long-term and short-term\\ memory using Conceptors}

\begin{document}

\maketitle

\begin{abstract}
We introduce a recurrent neural network model of working memory combining short-term and long-term components. The short-term component is modelled using a gated reservoir model that is trained to hold a value from an input stream when a gate signal is on. The long-term component is modelled using conceptors in order to store inner temporal patterns (that corresponds to values).  We combine these two components to obtain a model where information can go from long-term memory to short-term memory and vice-versa and we show how standard operations on conceptors allow to combine long-term memories and describe their effect on short-term memory.

\end{abstract}
\section{Introduction}

The reservoir computing (RC) paradigm \citep{Jaeger2004} is a peculiar and economic way to train a recurrent neural network (RNN) because only the output layer is modified while the input and recurrent layers are kept unmodified. Such RNNs are called \textit{reservoirs} because they provide a pool of non-linear computations based on inputs. Many variants (such as Echo State Networks \citep{Jaeger2001} and Liquid State Machine \citep{Maass2002}), along with specific extensions of this RC paradigm have been proposed since its initial stance by \citep{Jaeger2001} (for a review see \citep{Lukoeviius2009}), including implementations in various hardware like DNA- or laser-based ones (see \citep{Tanaka2019} for a recent review on physical reservoirs).
A recent and major enhancement of the RC paradigm has been proposed by \citet{Jaeger:2014}, called \textit{Conceptors}
(see Figure \ref{fig:conceptor} that introduces the main concepts). Intuitively, a conceptor represents a subspace of internal states of a RNN, e.g. the trajectory of a reservoir when fed by some input. This representation can later be used for extending the capacity of the original model. For instance it can be used to recognize temporal patterns \citep{Jaeger:2014, bao2016action, bartlett2019recognizing, Gast2017}, or to store and retrieve multiple arbitrary temporal patterns from a single RNN \citep{Jaeger:2014, Jaeger:2017}.  These conceptors have been recently used in a number of different works. For instance,  \citet{Mossakowskietal19} proposed an implementation of fuzzy logic based on conceptors,  \citet{DBLP:journals/corr/abs-1904-09187} used conceptors for online learning of sentence representations and \citet{he2018overcoming} proposed a general way to use conceptors during the learning of multiple tasks thanks to a reduction of interference between tasks by virtue of internal state space segregation. \\

Conceptors yield several advantages when compared to classical reservoirs, as Jaeger demonstrated it in his seminal paper \citep{Jaeger:2014}: conceptors provides symbolic operations on the latent space of the different input patterns. It is worth to be mentioned that such property were also exploited in the deep learning community and provided surprising results. For instance, in natural language processing (NLP), \citet{Mikolov2013} showed that arithmetic operations such as \enquote{$king - men + woman$} give a vector similar to \enquote{$queen$}. More recently, \citet{Brock2016} proposed a method to edit global image features based on operations performed on the latent space of generative adversarial networks (GANs). Conceptors provide similar logical operations but in the framework of the reservoir computing paradigm. For instance, \citet{Jaeger:2014} proposes an operator that quantifies if a stimulus is similar to an already known conceptor. By associating one conceptor per class, it is possible to measure if a stimulus belongs to a class (positive evidence) or none (negative evidence). Beyond logical operations, linear combinations of conceptors allow to implement continuous morphing between set of states: they were used to create morphing between two time series corresponding to the extended interpolation of the time series (e.g. a morphing between two sine-waves with different frequency is a sine-wave with an intermediate frequency).

Hypotheses on the interaction between working memory (WM) and long-term memory (LTM) has been explored in computational neuroscience models \citep{Nachstedt:2017}.
Moreover, the ability to store transient internal dynamics in long-term memory and to be able to retrieve theses dynamics later on when needed is an attractive principle for biology. It is to be related to some recent experimental observations suggesting that, in some specific cases\footnote{For instance, when some information to be maintained in working memory is known to be useful only later on.} \citep{Stokes:2015}, the maintenance of information is not uniform between the time of acquisition and the time of use. More specifically, it has been shown that the information cannot be reliably decoded from neural activity between these two times while it can be decoded at time of use. Some authors \citep{Mongillo:2008, Masse:2019, Manohar:2019} have thus suggested the existence of a mechanism to temporarily store information in synaptic weights (instead of being stored in neural activity). In this context, conceptors might provide a plausible explanation of such a transfer.

Based on the original RC paradigm, we have shown in \citep{Strock:2020} how a reservoir with feedback connections can implement a gated working memory, i.e. a generic mechanism to maintain information at a given time (corresponding to when the gate is on). This study shows that a reservoir, using a gating signal, can faithfully memorize triggered inputs for a while based on a stream of continuous values (i.e. working memory property). This model gives account on two important facts from neurosciences working memory (WM) studies: (1) the model is functionally a closed system when maintaining an information but it is physically an open system. This means the model can maintain information even when fed with a strong and continuous disturbing input. (2) Memory is encoded in the dynamics of the models and information can be maintained without trace of sustained activity.  In this model, the maintenance of information in output unit(s) is remarkably precise for long periods of time.

However, this model suffers from its cardinal property:
it is able to maintain any information with great precision because it is not backed up by a long term memory component; while in biology, what we hold in WM memory is directly influenced by our experience and perception.
Said differently, the current model is a quasi-perfect line-attractor, while it could be interesting to build a discrete line attractor model. For example, if we fed the model with a series of random scalar values with $10^{-3}$ precision, it may be desirable for the model to approximate (or discretize) the value to be maintained with only $10^{-1}$ precision. This might seem a trivial operation to be performed but it is actually harder than it seems because the robustness of the model is measured to the extent it is close to a formal line attractor. This is one of the reasons why we turned ourselves towards conceptors, to enable our WM model to be influenced by long-term memory.\\

In the present work, we introduce a link between short-term and long-term memory by combining two approaches: (1) a gated reservoir maintaining short-term information and (2) several conceptors maintaining long-term information. First, we introduce the transfer mechanisms between long-term and short-term memory. Then, we study how these mechanisms allow for a more stable version of the short-term memory and how it allows to add a prior to the values that must be maintained in short-term memory.  Finally, we explore the nature of operations carried on by conceptors, the different ways to combine memories and how this modifies short-term memory.

\section{Methods}

\subsection{Conceptors overview}

\begin{figure}[!ht]
    \begin{center}\textbf{How to load and retrieve one pattern ?}\end{center}
        \begin{center}
            \begin{minipage}{.4\linewidth}
                \begin{center}
                    \begin{minipage}{.5\linewidth}
                        \centering
                        N'\\
                        \includegraphics[align = c, width=\linewidth]{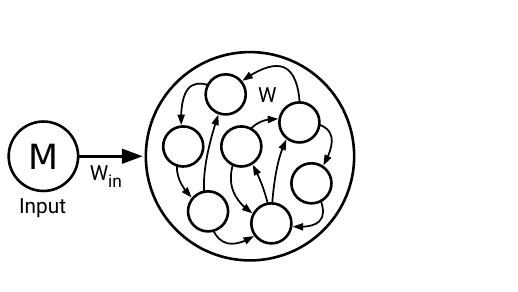}
                    \end{minipage}
                    {\hspace{-2em}\phantom{A}\large $\approx$\hspace{-1em}}
                    \begin{minipage}{.5\linewidth}
                        \centering
                        N\\
                        \includegraphics[align = c, width=\linewidth]{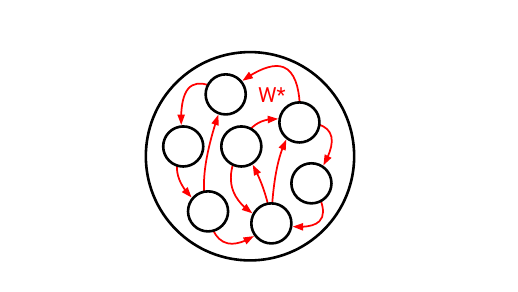}
                    \end{minipage}
                \end{center}
                \begin{center}
                    {\textbf{If}} N \textcolor{red}{mimic} N'\\
                    (Load)
                \end{center}
            \end{minipage}
            \begin{minipage}{.5\linewidth}
                \begin{center}
                    \begin{minipage}{.4\linewidth}
                        \centering
                        N'\\
                        \includegraphics[align = c, width=\linewidth]{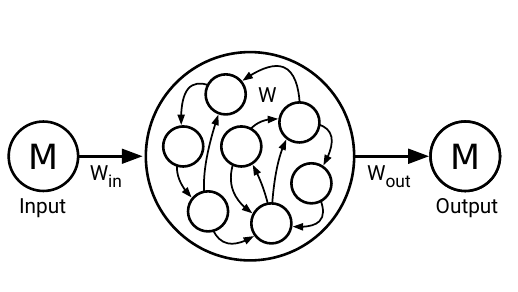}
                    \end{minipage}
                    {\large $\approx$\hspace{-1em}}
                    \begin{minipage}{.4\linewidth}
                        \centering
                        N\\
                        \includegraphics[align = c, width=\linewidth]{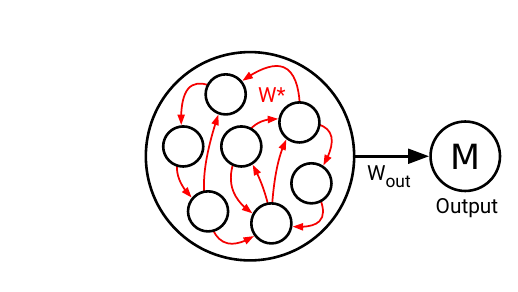}
                    \end{minipage}
                    \begin{center}
                        \hspace{-2em}
                        \textbf{then} M can be retrieved from N as it would be from N'\\
                        (Retrieve)
                    \end{center}
                \end{center}
            \end{minipage}
        \end{center}
    \begin{center}\textbf{How to load and retrieve several patterns ?}\end{center}
        \begin{center}
            \begin{minipage}{.4\linewidth}
                \begin{center}
                    \begin{minipage}{.5\linewidth}
                        \centering
                        N'\\
                        \includegraphics[align = c, width=\linewidth]{conceptor1.pdf}
                        \includegraphics[align = c, width=\linewidth]{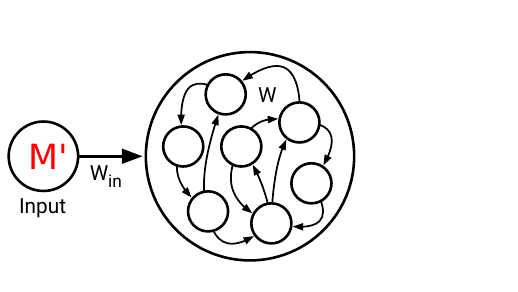}\\
                        N'
                    \end{minipage}
                    {\hspace{-2em}\large $\approx$\hspace{-1em}}
                    \begin{minipage}{.5\linewidth}
                        \centering
                        N\\
                        \includegraphics[align = c, width=\linewidth]{conceptor2-red.pdf}
                    \end{minipage}
                    \begin{center}
                        \hspace{-4em}
                        {\textbf{If}} N \textcolor{red}{mimic} N' when N' is receiving either M or \textcolor{red}{M'}\\
                        (Load)
                    \end{center}
                \end{center}
            \end{minipage}
            \begin{minipage}{.5\linewidth}
                \begin{center}
                    \vspace{3em}
                    \begin{minipage}{.4\linewidth}
                        \centering
                        N'\\
                        \includegraphics[align = c, width=\linewidth]{conceptor3.pdf}
                    \end{minipage}
                    {\large $\approx$\hspace{.5em}}
                    {\Large \textbf{?}}
                    \vspace{2.5em}
                    \begin{center}
                        \textbf{then} M can't be retrieved in the same way from N\\
                        \crossout{(Retrieve)}
                    \end{center}
                \end{center}
            \end{minipage}
        \end{center}
        \begin{center}
            \begin{minipage}{.4\linewidth}
                \begin{center}
                    \begin{minipage}{.5\linewidth}
                        \centering
                        N'\\
                        \includegraphics[align = c, width=\linewidth]{conceptor1.pdf}
                        \includegraphics[align = c, width=\linewidth]{conceptor1-2-red.pdf}\\
                        N'
                        \vspace{1em}
                    \end{minipage}
                    \hspace{-4em}
                    \includegraphics[align = c, width=.5\linewidth]{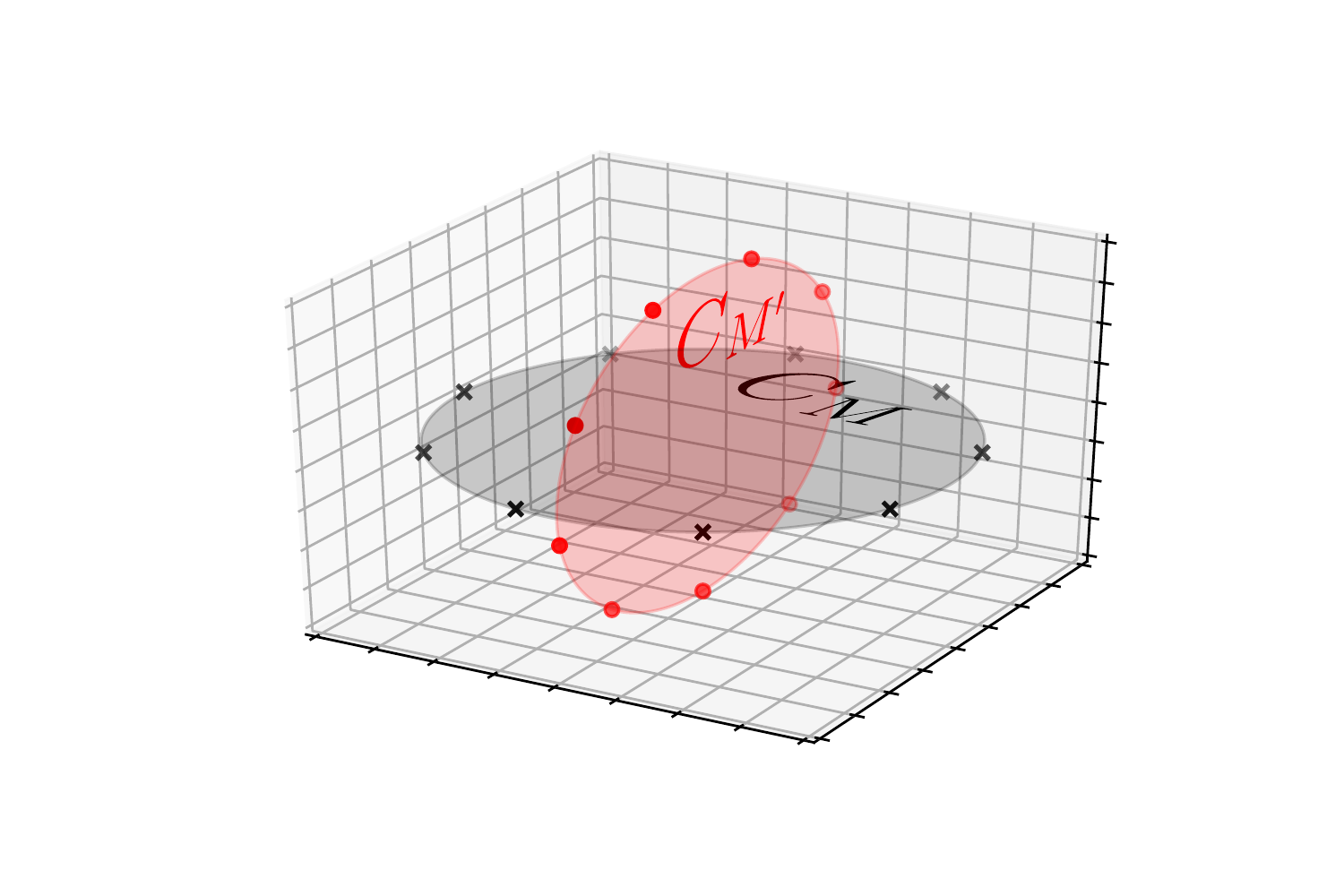}\\
                    \begin{minipage}{.8\linewidth}
                        \begin{center}
                            \hspace{-8em}
                            \textbf{But if} M and \textcolor{red}{M'} make evolve N' in  different spaces
                        \end{center}
                    \end{minipage}
                \end{center}
            \end{minipage}
            \begin{minipage}{.5\linewidth}
                \begin{center}
                    \begin{minipage}{.4\linewidth}
                        \centering
                        N'\\
                        \includegraphics[align = c, width=\linewidth]{conceptor1.pdf}
                    \end{minipage}
                    {\hspace{-2em}\large $\approx$\hspace{-1em}}
                    \begin{minipage}{.4\linewidth}
                        \centering
                        \hspace{2em}N\\
                        \vspace{.5em}
                        \includegraphics[align = c, width=\linewidth]{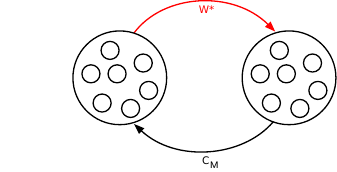}\\
                    \end{minipage}
                    \begin{center}
                        \textbf{then} N can \textcolor{red}{mimic} N' receiving M (or \textcolor{red}{M'}) by using an additional \textit{"projection"} $\text{C}_\text{M}$ (or \textcolor{red}{$\text{C}_\text{M'}$})
                    \end{center}
                    \begin{minipage}{.4\linewidth}
                        \centering
                        N'\\
                        \includegraphics[align = c, width=\linewidth]{conceptor3.pdf}
                    \end{minipage}
                    {\hspace{.5em}\large $\approx$\hspace{-1em}}
                    \begin{minipage}{.4\linewidth}
                        \centering
                        \hspace{2em}N\\
                        \vspace{.5em}
                        \includegraphics[align = c, width=\linewidth]{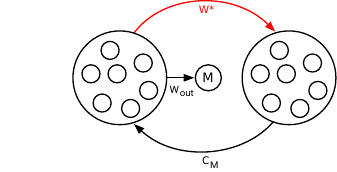}\\
                    \end{minipage}
                    \begin{center}
                        \hspace{-1em}\textbf{and} M (or \textcolor{red}{M'}) can be retrieved the same way from N\\
                        (Retrieve)
                    \end{center}
                \end{center}
            \end{minipage}
        \end{center}
    \caption{{\bf How to use conceptors to handle long-term memories of temporal patterns?} Let's assume $Ns$ are recurrent neural networks and $Ms$ are inputs or outputs. If N mimic N' then M can be retrieved from N as it would be from N'. If N mimic N' when N' is receiving either M or M' then M can't be retrieved the same way from N. But if M and M' make evolve N' in different spaces then N can mimic N' receiving M (or M') by using an additional \textit{"projection"} $\text{C}_\text{M}$ (or $\text{C}_\text{M'}$) and M (or M') can be retrieved the same way from N.
}
    \label{fig:conceptor}
\end{figure}

Compared to classical ESNs, conceptors represent a new training paradigm.
In Figure~\ref{fig:conceptor}, we show how conceptors can be used to store temporal patterns.
The training is performed in two steps and uses two models ($N$ and $N'$).
Let us consider a reservoir $N'$ that receives as input the sequence $M$ and that is trained (readout weights) to produce the sequence $M$ as output (auto-encoder). Let us now consider another reservoir $N$ that does not receive any input but is trained (internal weights) such as to match $N'$ internal activity (see figure \ref{fig:conceptor}). If we now read the internal activity of $N$ using the read-out weights of $N'$, we obtain the sequence $M$. Said differently, $N$ has learnt implicitly to spontaneously produce the sequence $M$. A very similar idea can actually be found in the full-FORCE training algorithm \citep{DePasquale:2018} where internal weights of a recurrent neural network are trained in order for the activity of its neurons to match the ones of another recurrent neural network that receives as input more information than the former one (e.g. the desired output)
Jaeger introduced a second idea based on the following observation: if a recurrent neural network is periodically stimulated with a sequence $M$, it evolves in a different region of space compared to when it is periodically stimulated with another sequence $M'$ (because we are in high dimensional space). Consequently, in order to have the possibility to generate two distinct patterns $M$ and $M'$, Jaeger proposes to train the model $N$ to match the activity of model $N'$ when it receives sequence $M$ or $M'$. In such a scenario, $N$ will end up following a mean trajectory, in between the trajectory where $N'$ receives $M$ as input and where $N'$ receives $M'$ as input. However, in light of the preliminary observation concerning the segregation of spaces for sequence $M$ and $M'$, it is possible to disentangle the activity within $N$ by projecting it to relevant sub-spaces. These sub-spaces can be identified in $N'$ when it receives $M$ or $M'$ respectively. A conceptor corresponds exactly to these projections.
In that context, these conceptors can be considered as long-term memories of the temporal patterns because they can be stored and reactivated later with negligible loss of recall/precision. More generally, conceptors can be considered as long-term memories of sets/subspaces of internal states.

\subsection{Models}

\subsubsection*{\bf Echo State Networks (ESN)}

In this work we consider Echo State Networks (ESN) with feedback from readout units to the reservoir. The system is described by the following update equations:
\begin{align}
  \label{eq:reservoir-conceptor}
  \begin{split}
    x[n] &= \tanh\left(W_{in} u[n]
                           + W      (x[n-1]+\xi)
                           + W_{fb} (y[n-1]) \right)\\
    y[n] &= W_{out} x[n]
  \end{split}
\end{align}
where $u[n]$, $x[n]$ and $y[n]$ are respectively the input, the reservoir and
the output at time $n$. $W$, $W_{in}$, $W_{fb}$, $W_{out}$ and $C$ are respectively
the recurrent, the input, the feedback, the output and the conceptor weight matrices and
$\xi$ is a uniform white noise term added to
reservoir units.

\subsubsection*{\bf Controlling ESN dynamics using a conceptor}

Following \citep{Jaeger:2014} notations, the equation for a conceptor $C$ enforcing some particular dynamics can be written as:
\begin{align*}
    x[n] &=  C\tanh\left(W x[n-1]+b\right)
\end{align*}
where $C$ is the conceptor (possibly changing over time), $x[n]$ is the state of the model  at time $n$, $W$ is the recurrent matrix and $b$ is a constant bias. This can be extended to the general case where we also have an input $u[n]$ (with input matrix $W_{in}$) (or similarly a feedback), and writes:
\begin{align*}
    x[n] &=  C\tanh\left(W x[n-1] +W_{in} u[n] + W_fb\right)\\
    y[n] &= W
\end{align*}

Using a conceptor $C$ is similar to a change of $W$ in $\tilde{W} = WC$ (and $W_\text{out}$ in $W_\text{out}C$ if there is feedback). In our implementation, we thus consider:
\begin{align}
    x[n] &= \tanh\left(WC x[n-1] +W_{in} u[n]\right)
    \label{eq:conceptor}
\end{align}

\subsubsection*{\bf Computing conceptors}
In order to compute a conceptor for some given dynamics, it is necessary to collect all the states of the reservoir and to concatenate them in a matrix $X$. The conceptor $C$ is then defined as:
\begin{align*}
    C = XX^T \left(XX^T+\alpha^{-2}I\right)^{-1}
      = R \left(R+\alpha^{-2}I\right)^{-1}
\end{align*}
where $R = X X^T$ is similar to a covariance matrix, and $\alpha$ (a.k.a the aperture) controls how close from the identity matrix $C$ is.

\subsubsection*{\bf Aperture adaptation}
Intuitively, the aperture of a conceptor controls the precision of the internal states representation.
However, no information on internal states is lost, because it is possible to change the aperture of a conceptor $C$ without the need to recompute the conceptor from scratch.
To change the aperture, one only need to adapt the conceptor $C$ as follows:
\begin{align}
    \phi(C, \gamma) &= C\left(C+\gamma^{-2}(I-C)\right)^{-1}
\end{align}
where $\phi(C, \gamma)$ represents the same states than $C$ with a different aperture, and $\gamma$ is controlling how the aperture is modified.
Intuitively, $\phi(C, \gamma)$ modifies the aperture of $C$ by a factor of $\gamma$.

\subsubsection*{\bf Linear combination}
Given two conceptors $C_1$ and $C_2$ and $\lambda\in\mathbb{R}$, the linear combination of conceptor $C_1$ and $C_2$ is defined as:
\begin{align*}
C &= \lambda C_1 + (1-\lambda) C_2
\end{align*}
In the following when $\lambda\in [0,1]$ we will talk about interpolation, when $\lambda > 1$ about right-extrapolation, and when $\lambda < 0$ about left-extrapolation.

\subsubsection*{\bf Boolean operations}
Boolean operations can be written as:
\begin{align*}
C\vee B &= \left(I+\left(C(I-C)^{-1}+B(I-B)^{-1}\right)^{-1}\right)^{-1}\\
C\wedge B &= \left(C^{-1}+B^{-1}-I\right)^{-1}\\
\neg C &= I-C
\end{align*}
However, as highlighted in \citep{Mossakowski:2019}, $\vee$ and $\wedge$ are not indempotant (i.e. $C\vee C \neq C$ and $C\wedge C \neq C$). More precisely if $C$ (resp. $B$) is a conceptor built with the covariance matrix $R$ (resp. $Q$), Jaeger proposes to build $C\vee B$ using the covariance matrix $R+Q$ that is by design not indempotent.
What we propose here is to consider instead the matrix $\beta R+(1-\beta)Q$ with $\beta\in [0, 1]$ instead of $R+Q$, or if we want it to be symmetric $(R+Q)/2$.
Similar calculation gives the following new $\vee_\beta$ and $\wedge_\beta$.
\begin{align*}
C\vee_\beta B &= \left(I+\left(\beta C(I-C)^{-1}+ (1-\beta)B(I-B)^{-1}\right)^{-1}\right)^{-1}\\
C\wedge_\beta B &= \left(\beta C^{-1}+ (1-\beta) B^{-1}\right)^{-1}
\end{align*}
This way of building the OR operation also has a data driven intuition. If we note $\beta = \frac{n}{n+p}$ where $n$ (resp. $p$) is the number of data points used to build $R$ (resp. $Q$) then $bR+(1-b)Q$ is the "correlation matrix" obtained by taking the union of all the data points.
Moreover, if we choose $\beta = 0.5$ then there is a direct link between the two way of defining the OR: $C\vee B = \phi(C\vee_{0.5}B, 2)$.
In this study, the aperture was mostly not influencing the results, thus we show only the results for $\vee$.

\subsection{Tasks}

\begin{figure}[ht]
    \centering
    \includegraphics[width = \textwidth]{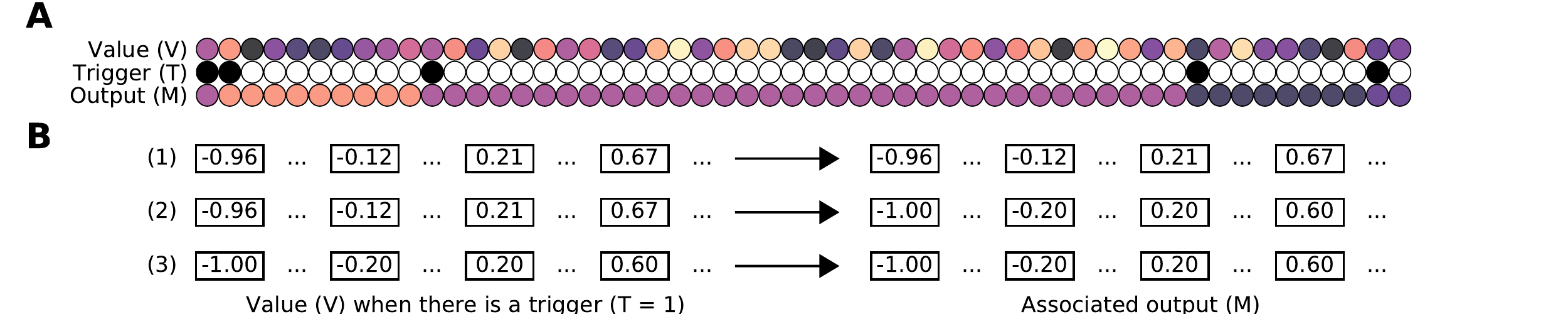}
    \caption{Gating tasks and its two variants. \textbf{A.} The gating task. Each column represents a time step (time increases from left to right), colored discs represent inputs ($V$ and $T$) and the output ($M$).
    \textbf{B.}~Variant tasks: \textbf{(1)}~Original task: from continuous to continuous values. \textbf{(2)}~First variant (C2D task): from Continuous to Discrete values. \textbf{(3)}~Second variant (D2D task): from Discrete to Discrete values.}
    \label{fig:tasks}
\end{figure}

We consider the gating task described in \citep{Strock:2020} and two additional variants. In this task the model receives an input $V$ that is continuously varying over time and another input being either 0 or 1 (trigger or gate $T$). To complete the task, the output has to be updated to the value of the input when the trigger is active and to remain constant otherwise (similarly to a line attractor). In other words, the trigger acts as a gate that controls the entry of the value in the memory (the output). Figure \ref{fig:tasks} describes this task and the two variants we consider in this work. In both variants we consider 11 values uniformly spread between -1 and 1: these values are used to discretize the input value (V) (when a trigger occurs; T=1) stored in the associated output (M). In the first variant (C2D task) we discretize only the associated output whereas in the second variant (D2D task) both are discretized. On a concrete example, if the model was trained to maintain 0.41 ($M=0.41$), that means it was receiving a trigger ($T=1$) along with the value 0.41 ($V=0.41$). In the first approach we change both $V$ and $M$ to $0.4$, whereas in the second approach we change only $M$ to $0.4$.

\subsection{Implementation details}

We consider a reservoir of 1000 neurons that has been trained to solve a gating task described in \citep{Strock:2020}. The overall dynamics of the network we consider are described by the following equations:
\begin{align}
  \label{eq:reservoir-conceptor}
  \begin{split}
    x[n] &= C\tanh\left(W_{in} u[n]
                           + W      (x[n-1]+\xi)
                           + W_{fb} (y[n-1]) \right)\\
    y[n] &= W_{out} x[n]
  \end{split}
\end{align}
where $u[n]$, $x[n]$ and $y[n]$ are respectively the input, the reservoir and
the output at time $n$. $W$, $W_{in}$, $W_{fb}$, $W_{out}$ and $C$ are respectively
the recurrent, the input, the feedback, the output and the conceptor weight matrices and
$\xi$ is a uniform white noise term added to
reservoir units.
$W$, $W_{in}$, $W_{fb}$ are uniformly sampled between $-1$ and $1$.
Only $W$ is modified to have sparsity level equal to $0.5$ and a spectral radius of $0.1$.
When $W_{out}$ is computed to solve the gating task, the conceptor $C$ is considered to be fixed and equal to the identity matrix ($C=I$).
In normal mode, the conceptor $C$ is equal to a conceptor $C_m$ that is generated and associated to a constant value $m$.
In order to compute this conceptor $C_m$, we impose a trigger ($T=1$) as well as the input value ($V=m$) at the first time step, such that the reservoir has to maintain this value for 100 time steps. During these 100 time steps, we use the identity matrix in place of the conceptor. The conceptor $C_m$ is then computed according to $C_m = XX^T \left(XX^T+\frac{I}{a}\right)^{-1}$, where $X$ corresponds to the concatenation of all the 100 reservoir states after the trigger, each row corresponding to a time step, $I$ the identity matrix and $a$ the aperture. In all the experiments the aperture has been fixed to $a=10$. For the 
conceptors pre-computed in Figure \ref{fig:approximation} and \ref{fig:generalisation}, the reservoir have been initialised with its last training state.

\section{Results}

\subsection*{Transfer between long-term and short-term memory}
\begin{figure}[ht]
    \centering
    \includegraphics[width = \textwidth]{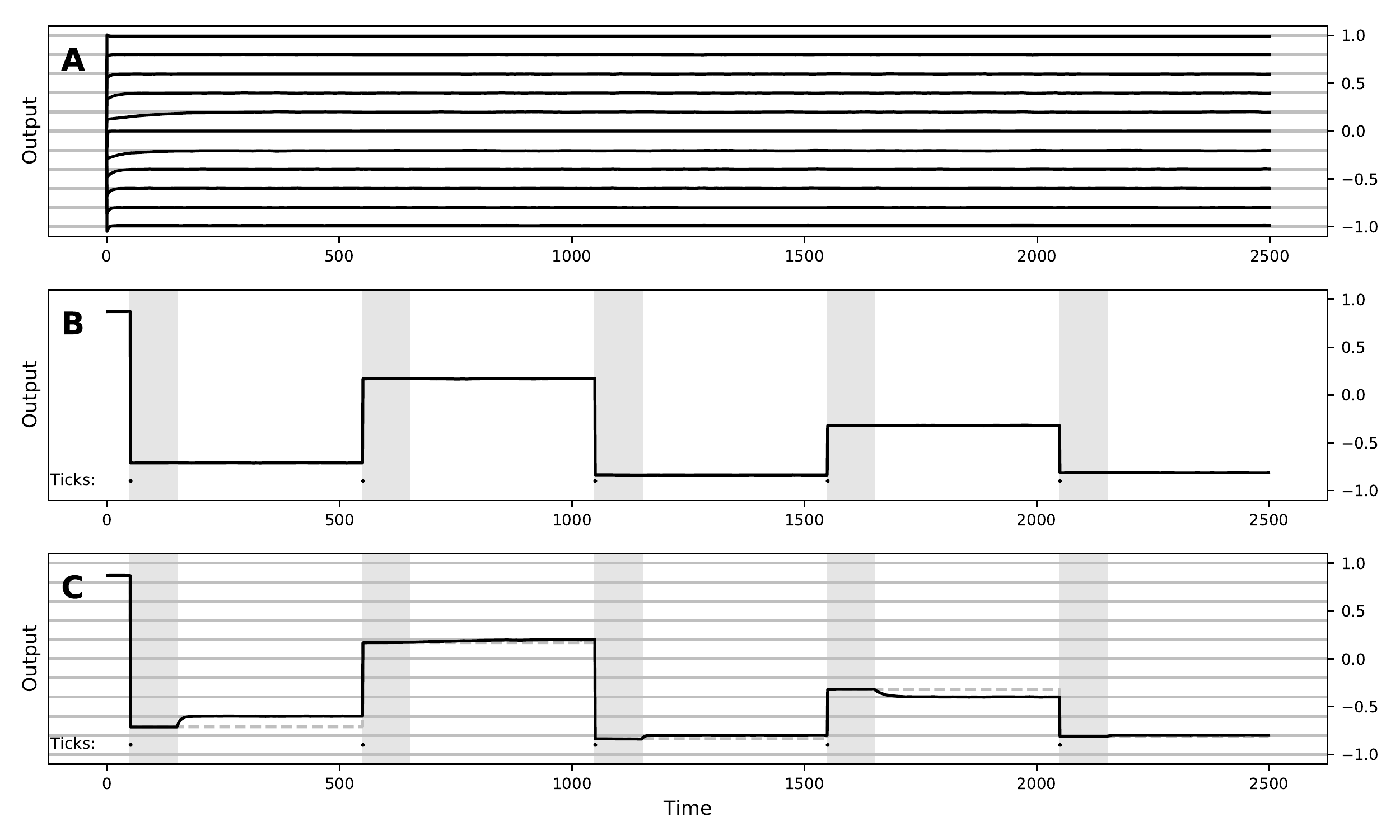}
    \caption{Approximation with conceptors or discrete conceptors. Black: Evolution of the short-term memory (i.e. the readout $y$). Gray lines: the discrete value considered. Light gray areas: time when conceptors are computed for the current value stored in short-term memory. \textbf{A.} Discrete conceptors are appplied from a random state. \textbf{B}-\textbf{C.} Conceptor $C_m$ are computed using 100 time steps after a trigger while $C = I$. \textbf{B.} $C_m$ is directly applied to the following time steps. \textbf{C.} The closest conceptor among the discretized conceptor is applied for the following time steps. Dashed lines represents the memory that should have been kept if not discretized.}
    \label{fig:approximation}
\end{figure}

Figure \ref{fig:approximation} displays the two core ideas of our approach: \textbf{(1)} How to transfer short-term to long-term memory and \textbf{(2)} How to retrieve (in short-term memory) an information stored in long-term memory.
\textbf{(1)} The long-term memory we consider is the conceptor $C_m$ associated to the value $m$ maintained in short-term memory. To compute $C_m$ we use the 100 first time steps after a trigger. Meanwhile no conceptor is applied (i.e. $C = I$). After that we update $C$ with $C_m$. On figure \ref{fig:approximation}B we can see that it doesn't seem to cause any interference in the short-term memory. However, the memory currently lies both in the conceptor $C$ (long-term) and in the output unit $y$ (short-term).
\textbf{(2)} Now, the long-term memory we consider are only conceptors $C_m^D$ associated to discrete values between -1 and 1 (11 values uniformly spread between -1 and 1). Similarly as before, after a trigger we compute a new conceptor $C_m$ using the 100 first time steps after a trigger and without conceptors ($C = I$). Then, we search for the closest conceptor $C_m^i$ among the conceptors with discrete values $C_m^D$ using a distance between conceptors and we update $C$ with this conceptor. On figure~\ref{fig:approximation}C, we see the following behavior: after a trigger, the value is correctly updated in short-term memory and remains stable until $C$ is updated (after 100 time steps) and then the output jumps to the closest discrete representation of the memory.\\

\subsection*{Extended maintenance (well beyond learning)}

\begin{figure}
    \centering
    \includegraphics[width=\linewidth]{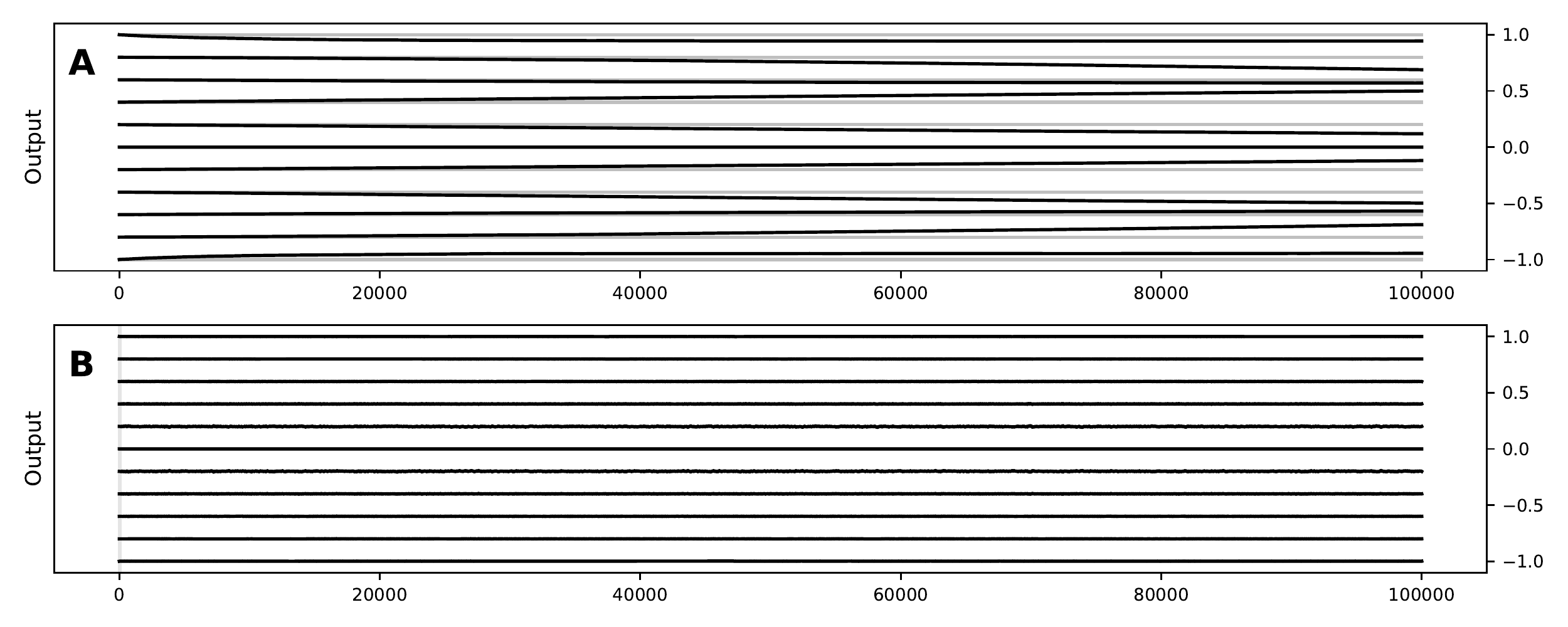}
    \includegraphics[width=\linewidth]{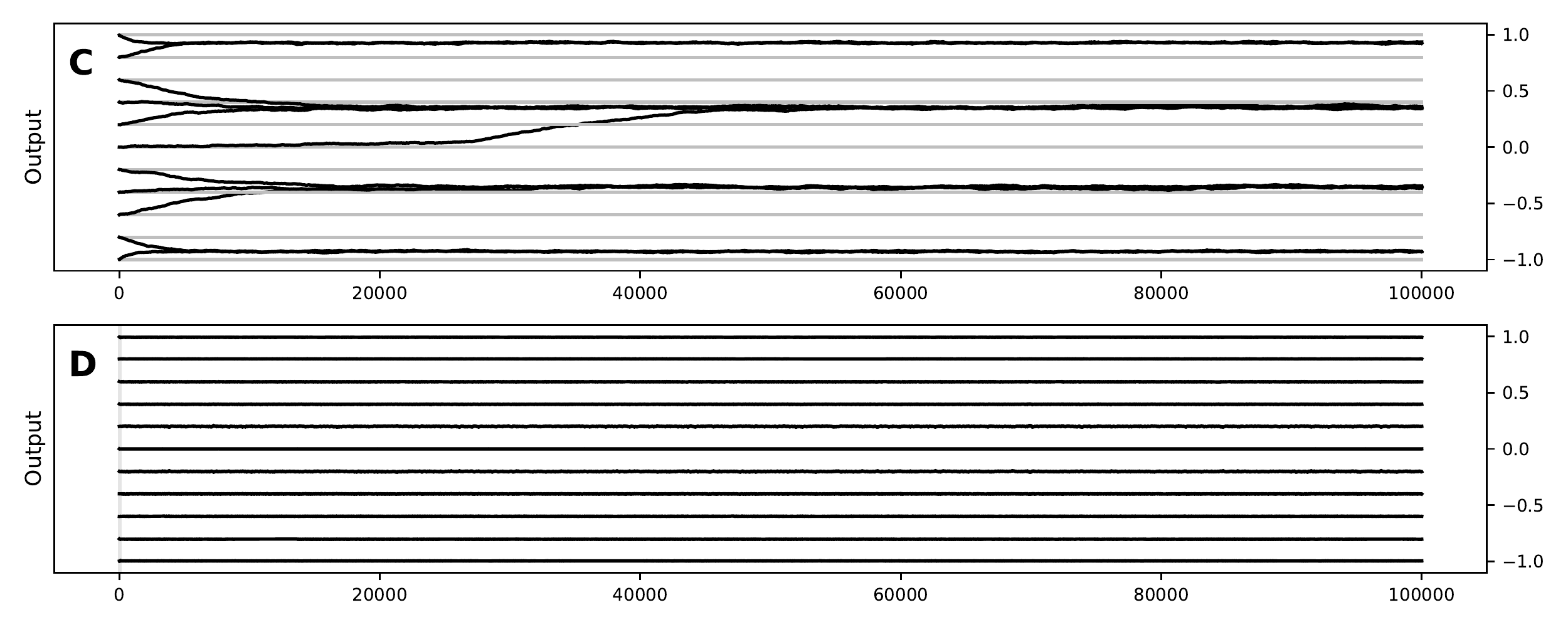}
    \caption{Stability comparison with or without conceptor, with or without noise. Black: Evolution of the short-term memory (i.e. the readout $y$). Gray lines: the discrete value considered. Light gray areas: time when conceptors are computed for the current value stored in short-term memory.
    \textbf{A}-\textbf{B} No noise.
    \textbf{C}-\textbf{D} 0.0001 noise.
    \textbf{A} and \textbf{C} No conceptor used.
    \textbf{B} and \textbf{D} Discrete conceptor are applied.}
    \label{fig:stability}
\end{figure}

In Figure~\ref{fig:stability}, we show how conceptors allow to better stabilize the short-term memory, even in the presence of noise.
Let us start by reminding that both the model and the conceptors have been trained to maintain information in short-term memory only for few hundreds of time step, and that in the constant presence of a disturbing input (V).
However, in the absence of noise ($\xi$ = 0), we can clearly see that even without conceptors the short-term memory can be maintained for several thousands of time steps.
Nevertheless the ability to maintain information in short-term memory is not infinite: if we go further in time we can note that after approximately 100,000 time steps the short-term memory will slowly degrade (Figure~\ref{fig:stability}A). The first thing noticeable is that, with conceptors, this slow degradation vanishes (Figure~\ref{fig:stability}B). Moreover, we tried the same analysis with noise inside the reservoir.
Even $10^{-4}$ noise ($std(\xi) = 10^{-4}$ ) prevents the model to maintain longer than it has been trained to (Figure~\ref{fig:stability}C). Interestingly in the noisy case, the benefit of conceptors is even more visible since they allow to maintain information in memory as if there were no noise (Figure~\ref{fig:stability}D).

\subsection*{Using constant-memory conceptor stabilizes faster and more accurately to discretized values.}
\begin{figure}
    \centering
    \includegraphics[width = \linewidth]{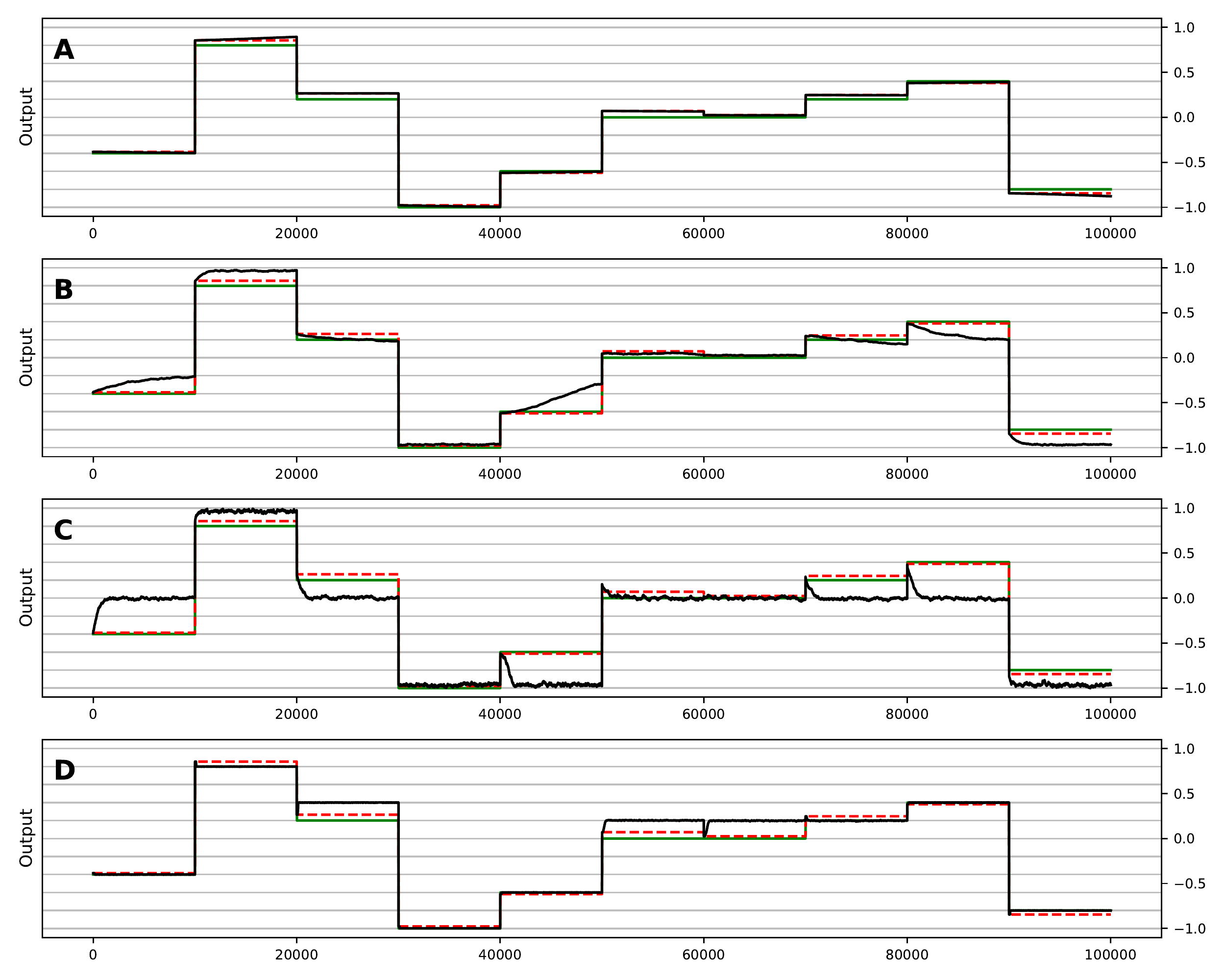}
    \caption{\textbf{Different training procedures to discretize the output.} \textbf{Black lines}: Evolution of the short-term memory (i.e. the readout $y$).
    \textbf{Gray lines}: the discrete value considered.
    \textbf{Green lines} represent the memory that should be kept when discretized.
    \textbf{Red dashed lines} represent the memory that should have been kept if not discretized.
    \textbf{A} \textbf{D2D task}: Only discrete values have been trained to be maintained.
    \textbf{B}-\textbf{C} \textbf{C2D task}: The value maintained is a discretized version of the value coming along with the trigger and no more the value itself.
    \textbf{B} Offline learning. \textbf{C} Online learning (ridge = 0.0001).
    \textbf{D} \textbf{Constant-memory conceptor for discretized values}: Conceptors are computed using the 100 time steps after a trigger while $C = I$ and then the closest conceptor among the discretized conceptor is applied for the following time steps.
    The period of time during which conceptors are computed cannot be shown because they are too short compared to the figure's timescale.
    }
    \label{fig:discretization}
\end{figure}

Another benefit of augmenting the WM model with conceptors is that while keeping the ability to maintain possibly everything, it is possible to add priors on the values the model is more likely to maintain. In Figure~\ref{fig:discretization}, we show the outcome of two complementary approaches discretizing memory.  As we had already noticed in \citep{Strock:2020}, only by being trained to maintained few discrete memory the model seems to generalize to all real values between -1 and 1. Thus the first task (C2D task) does not seem to allow to discretize the memory (Figure~\ref{fig:discretization}A). Concerning the second task (D2D task), it seems able to discretize the value memorized, (offline training Figure~\ref{fig:discretization}B, online training Figure~\ref{fig:discretization}C). But, in case it converges, the convergence towards a discrete value is way slower than with conceptors (Figure~\ref{fig:discretization}D).

\subsection*{Aperture adaptation of a constant-memory conceptors}
The first operation we considered on constant-memory conceptors was the aperture adaptation. We did not notice an apparent influence of the aperture on the ability to maintain the constant values as long as it is not too small.

\subsection*{Linear interpolation of two constant-memory conceptors}
\begin{figure}[ht]
    \centering
    \includegraphics[width = \textwidth]{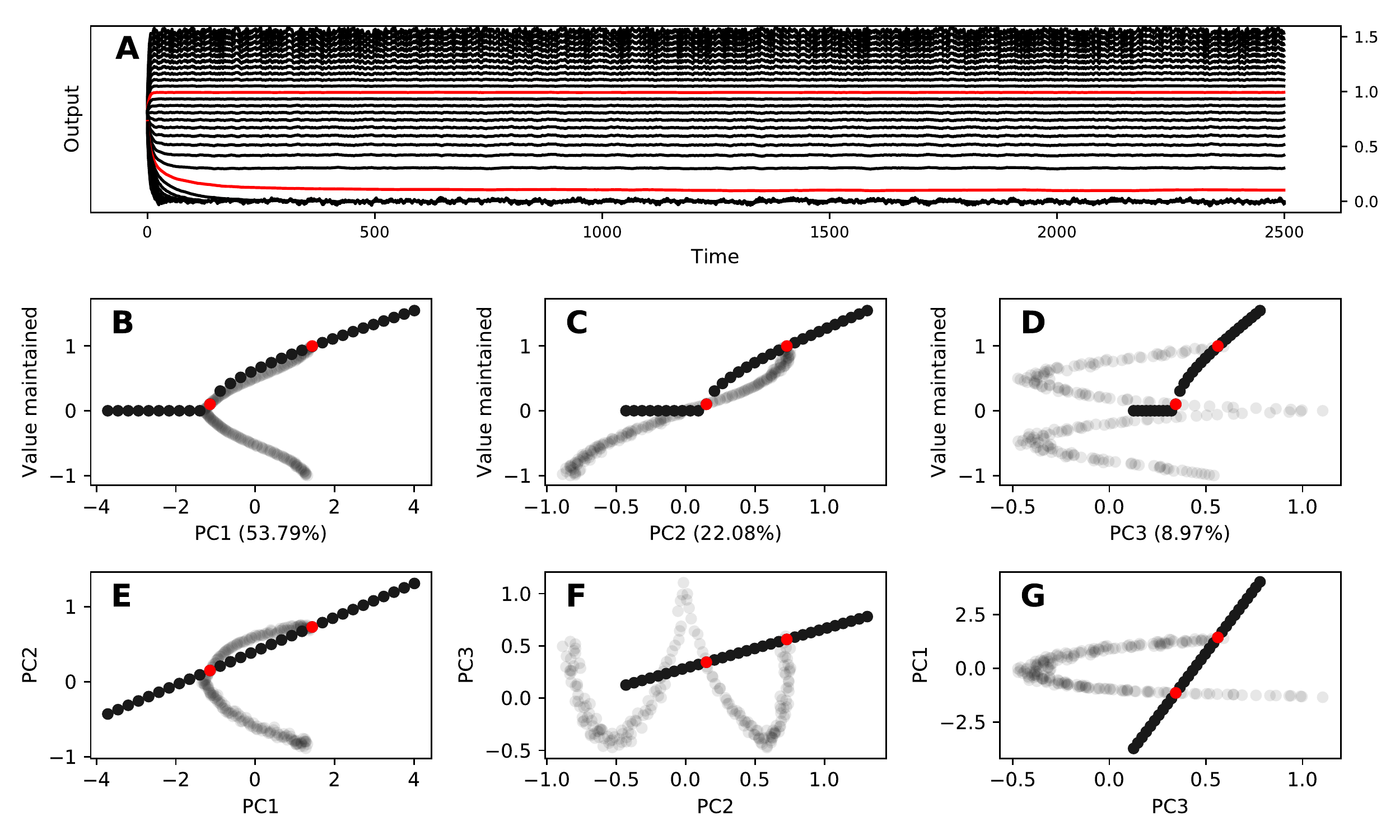}
    \caption{Generalisation of constant-memory conceptors $C_m$. \textbf{Red}: two constant-memory conceptors: $C_{0.1}$ and $C_{1.0}$. \textbf{Black}: Inferred conceptors, i.e. linear interpolation and extrapolation between $C_{0.1}$ and $C_{1.0}$. \textbf{A} Evolution of the short-term memory against time for different conceptors. \textbf{B}-\textbf{G} \textbf{Gray}: constant-memory conceptors $C_{m}$ for 201 values of $m$ uniformly spread between $-1$ and $1$. \textbf{B}-\textbf{D} Link between principal components of the conceptors and the memory they are encoding. For the interpolated conceptors, the memory is considered as the mean in the last 1000 time steps. \textbf{E}-\textbf{G} Representation of the conceptors in the three principal components of the $C_m$ conceptors.}
    \label{fig:generalisation}
\end{figure}

In Figure \ref{fig:generalisation}, we show two main ideas: \textbf{(1)} how a linear interpolation between two conceptors can allow to generalize the gating of other values, and \textbf{(2)} a representation of the space in which lies the conceptors and their link to the memory they encode.
\textbf{(1)} Interpolation and extrapolation $C$ of conceptor $C_{0.1}$ and conceptor $C_{1.0}$ has been computed as $C = \lambda C_{1.0} + (1-\lambda) C_{0.1}$ with 31 $\lambda$ values uniformly spread between -1 and 2.
Even though the interpolated ($\lambda\in[0, 1]$) conceptors obtained are not exactly equivalent to $C_m$  conceptors obtained in Figure~\ref{fig:approximation}, they seem to also correspond to a retrieved long-term memory value to be maintained.
The mapping between $\lambda$ and the value is non-linearly encoded. For right-extrapolation ($\lambda\in[1, 2]$) the conceptor seems to be linked to a noisy version of a $C_m$ conceptor. A value seems still to be retrieved from long-term memory and maintained in short-term memory: the output activity is not constant, but its moving average is constant. For left-extrapolation ($\lambda\in[-1, 0]$), the conceptor obtained does not seem to encode any information anymore: all the output activities collapse to zero.
\textbf{(2)} Principal Component Analysis (PCA) have been performed using 201 pre-computed conceptors associated to values uniformly spread between -1 and 1. The first three components already explain approximately 85\% of the variance. The first component seems to non-linearly encode the absolute value of the memory (Figure \ref{fig:generalisation}B) whereas the second component seems to non-linearly encode the memory itself (Figure \ref{fig:generalisation}C). The straight line conceptor $C_m$   (Figure \ref{fig:generalisation}E-G), that might be an explanation why extrapolation does not work as we expected.

\subsection*{Intersection and union of constant-memory conceptors}
\begin{figure}[ht]
    \centering
    \includegraphics[width = \textwidth]{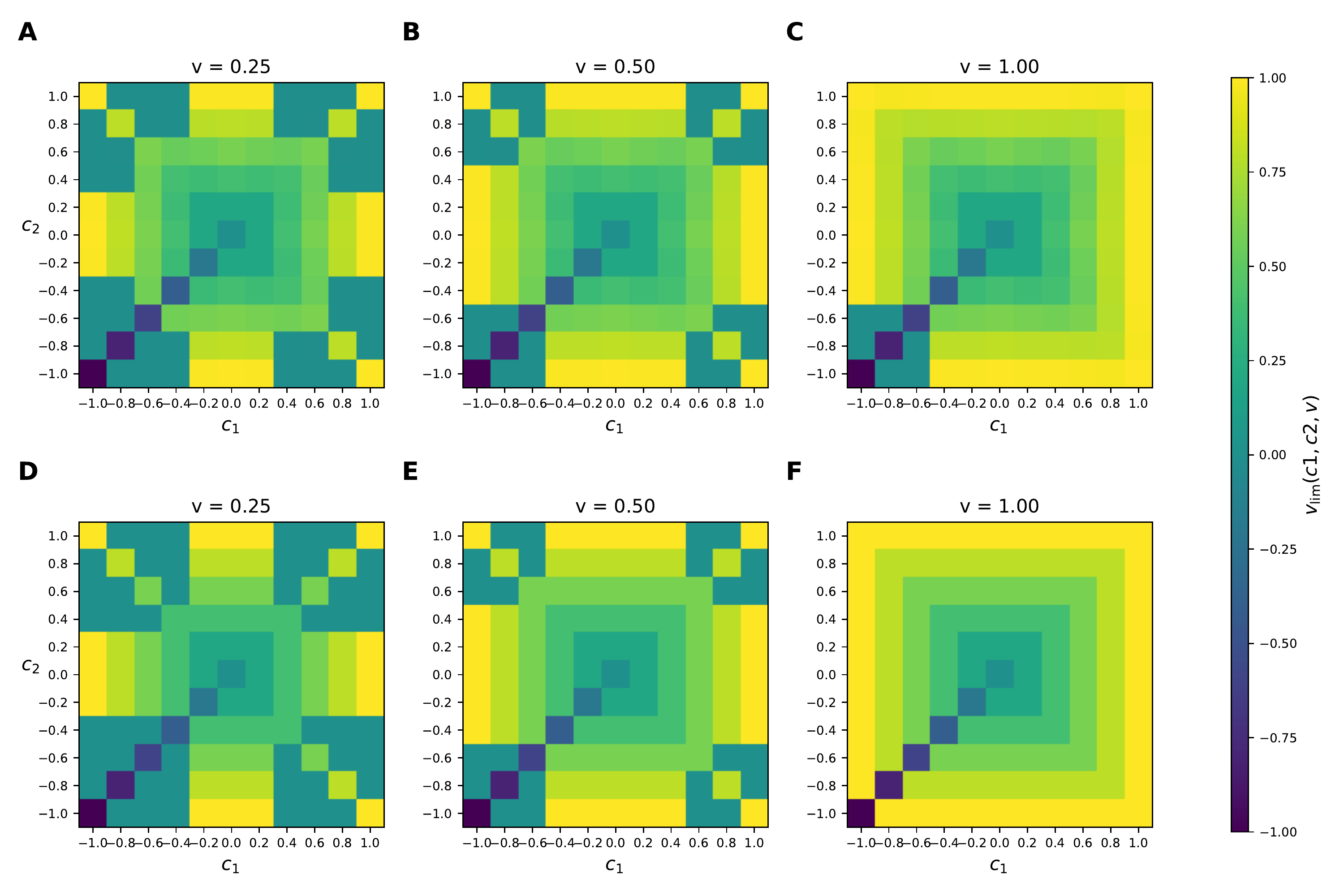}
    \caption{\textbf{Relaxation values} (i.e. values towards which the output relaxes) when applying the disjunction of two constant-memory conceptors.
    In other words, it corresponds to the final values reached when applying the conceptor $C_1\vee C_2$.
    \textbf{A}-\textbf{C} Empirical results from experiments.
    \textbf{D}-\textbf{F} Predictions based on equation \ref{eq:formula}.
    }
    \label{fig:disjunction}
\end{figure}

We studied both how the conjunction and the disjunction of constant-memory conceptors were influencing the dynamics.
In both cases, when a trigger occurs the output jumps towards the value to be maintained and then relaxes to another value.
In the conjunction case, the value towards which it relaxes is easy to describe, it is always almost zero.
In the disjunction case, it is harder to describe.
In Figure~\ref{fig:disjunction}, we show the values towards which the output relaxes (i.e. relaxation values) when the disjunction of two constant-memory conceptors is applied.x
First, as the disjunction of twice the same conceptor is either the same conceptor or an aperture adaptation of it (i.e. $C_m\vee C_m = \phi(C_m,2)$ and $C_m\vee_\beta C_m = C_m$), the value towards which it relaxes is the value of the conceptor itself.
Then, we realized that we could predict what would be the relaxation values in different cases: in general the relaxation value was mostly either almost zero or the maximum of the absolute values of the two conceptors multiplied by the sign of the new value to maintain.
We propose the following formula to predict the value towards which it relaxes:
\begin{align}
v_{\lim}(c_1,c_2,v) &=
    \begin{cases}
        c_1 &\text{if } c_1 = c_2 \\
        sign(v)\times max(|c_1|, |c_2|) &\text{if } min(|c_1|, |c_2|)<|v| \text{ or } c_1 = -c_2\\
        0 &\text{otherwise}
    \end{cases}
    \label{eq:formula}
\end{align}
where $v$ is the initial value ($V$) proposed along with the trigger,
$c_1$ (resp. $c_2$) is the constant associated to conceptor $C_1$ (resp. $C_2$),
$v_{\lim}(c_1,c_2,v)$ is the ultimate value reached while applying conceptor $C_1\vee C_2$.

The predictions made by the formula are less accurate for extreme values such as for $v = 1.00$ (see Figure~\ref{fig:disjunction}).
We hypothesize a similar formula for relaxation values of $n$ constant-memory conceptors:
\begin{align}
    v_{\lim}(c_1,...,c_n,v) &=
    \begin{cases}
        c_1 &\text{if } c_1 = c_2 = ... = c_n \\
        sign(v)\times max(|c_1|,|c_2|, ...,|c_n|) &\text{if } min(|c_1|, |c_2|, ..., |c_n|)<|v|\\
        & \text{or } (\forall i,j\ |c_i|=|c_j|\\
        & \text{and } \exists i,j\ \text{such that } i>j \text{ and } c_i = -c_j)\\
        0 &\text{otherwise}
    \end{cases}
\end{align}
where $v$ is the initial value ($V$) proposed along with the trigger,
$c_i$ is the constant associated to conceptor $C_i$,
$v_{\lim}(c_1,c_2,...,c_n,v)$ is the ultimate value reached while applying conceptor $\bigvee\limits_{i = 1}^n C_i$.

\section{Discussion}

This study introduces the basis for establishing the link between long-term and short-term working memory in echo state networks using conceptors. This allowed us to show how a short-term working memory can be formed under the influence of long-term memory: i.e. how working memory can be biased towards predefined discrete values stored in long-term memory. In most working memory models (e.g. \cite{Lim2013, Bouchacourt2019}), it is implicitly assumed that the working memory is faithful to the value(s) to be maintained, that is, without any influence from the long-term memory. However, biological and psychological observations suggest that working memory is influenced by long-term memory.
First, perceptions may be perturbed by noisy inputs or other processes, thus the associated working memory might ends up being different from the ground truth. Second, our past experience may influence the information that will be maintained. This is exactly what we have shown using conceptors and an ad-hoc method for updating the working memory. Future work will concentrate on removing the (less biologically plausible) engineered steps, namely, the offline computation and selection of the closest conceptor. Such processes could be implemented using the auto-conceptors introduced in \citep{Jaeger:2014}.

There are however theoretical difficulties when combining conceptors together: the result is difficult to predict because it largely differs from what we would naturally expect. For instance, a linear interpolation of constant-memory conceptors does not create another constant-memory conceptor. The reason being that the space of constant-memory conceptors is not a straight line. Hence, a mere linear combination of constant-memory conceptors could not lead to another constant-memory conceptor. Nevertheless, we have shown empirically that in all scenarios a linear combination of two constant-memory conceptors lead to a value that is maintained. However, this new memory is oscillating around the combination of the constant values (see Figure~\ref{fig:generalisation}). This oscillation being a direct consequence of the perturbation of the system (i.e. the input). Moreover, the disjunction of conceptors is not implementing what we were expecting. 
For two conceptors with two constant values $v_1$ and $v_2$ such that $0 \leq v_1 \leq v_2$, we would expect that the disjunction encodes the two values simultaneously. More specifically, we expected such disjunction to implement a choice function between the two values stored in long-term memory.
Instead, we obtained a conceptor that does not converge towards $v_1$ but only towards $0$ or $v_2$ depending on the given input value. To some extent, $v_1$ and $v_2$ influence the disjunction with however different qualitative roles. In the general case of a disjunction of $n > 2$ constant-memory conceptors, only the extreme value seems to matter in the composite conceptor.

More interestingly, this work opens the door to another form of working memory: procedural (or functional) working memory. Instead of temporarily memorizing declarative information, this kind of working memory would be able to memorize procedural information (e.g. how a task should be performed, which processes should be applied, etc.). For instance, imagine you are given some instructions which are to sum up a series of numbers. In order to complete this task, it is necessary to keep track of the current sum (e.g. in a classical short-term declarative working memory) that needs to be updated each time a new number is given. However, it is also necessary to remember the preliminary instruction (i.e summing up) in another form of working memory which is long-term (it needs to span the whole experiment) and which is procedural.
This procedural nature makes this working memory quite peculiar because instead of memorizing a given information, it needs to memorizes a procedure -- here, a sequence of operations depending on the context -- that needs to be applied each time an input is given. It is not yet clear how such memory could be encoded in the brain (e.g. sustained activity, dynamic activity, transient weights) and we think conceptors might be key in answering this question, but more experimental and theoretical work will be needed before answering this question.

\section*{Compliance with Ethical Standards}
Authors declare they have no conflict of interest.
Ethical approval: This article does not contain any studies with human participants or animals performed by any of the authors.






\end{document}